\documentclass{article}

\usepackage[preprint]{neurips_2025}

\usepackage[utf8]{inputenc} 
\usepackage[T1]{fontenc}    
\usepackage{hyperref}       
\usepackage{url}            
\usepackage{booktabs}       
\usepackage{amsfonts}       
\usepackage{nicefrac}       
\usepackage{microtype}      
\usepackage{xcolor}         
\usepackage{amsmath}
\usepackage{tabularx}
\usepackage[most]{tcolorbox}
\tcbuselibrary{breakable}
\definecolor{lightgreen}{rgb}{0.4,0.85,0.4}

\usepackage{caption}
\usepackage{subcaption}
\usepackage{bbding}
\usepackage{wrapfig}
\usepackage{graphicx}
\usepackage{multirow}
\usepackage{tabu}
\usepackage{wrapfig}

\AtEndPreamble{
    \usepackage[capitalize]{cleveref}
    \crefname{section}{Sec.}{Secs.}
    \Crefname{section}{Section}{Sections}
    \Crefname{table}{Table}{Tables}
    \crefname{table}{Tab.}{Tabs.}
}
\usepackage{nicematrix}
\usepackage{colortbl}
\definecolor{mygray}{gray}{0.95}
\definecolor{my_green}{RGB}{82,208,80}
\usepackage{enumerate}
\usepackage{makecell}
\definecolor{00red}{RGB}{236,35,35}
\newcommand{\whred}[1]{\textcolor{00red}{#1}}
\usepackage{arydshln} 
\usepackage{pifont}
\definecolor{Highlight}{rgb}{0.92,0.94,1}

\usepackage{subcaption} 

\usepackage{enumitem}
\usepackage{xspace}
\newcommand{\methodname}{R1-Compress\xspace}

\title{R1-Compress: Long Chain-of-Thought Compression via Chunk Compression and Search}

\author{%
  Yibo Wang$^{1}$\footnotemark[1],
  Haotian Luo$^{2}$\footnotemark[1],
  Huanjin Yao$^{1}$,
  Tiansheng Huang,
  Haiying He$^{2}$\\
  \textbf{Rui Liu}$^{3}$,
  \textbf{Naiqiang Tan}$^{3}$,
  \textbf{Jiaxing Huang}$^{4}$,
  \textbf{Xiaochun Cao}$^{2}$
  \textbf{Dacheng Tao}$^{4}$ 
  \textbf{Li Shen}$^2$\footnotemark[2]
  \\
  $^1$ Tsinghua University \quad
  $^2$ Shenzhen Campus of Sun Yat-sen University \\
  $^3$ Didichuxing Co. Ltd \quad
  $^4$ Nanyang Technological University \\
}

\begin{document}

\renewcommand{\thefootnote}{\fnsymbol{footnote}}
\footnotetext[1]{Equal contribution}
\footnotetext[2]{Corresponding Author: Li Shen (shenli6@mail.sysu.edu.cn)}
\renewcommand{\thefootnote}{\arabic{footnote}}

\maketitle

\begin{abstract}
Chain-of-Thought (CoT) reasoning enhances large language models (LLMs) by enabling step-by-step problem-solving, yet its extension to Long-CoT introduces substantial computational overhead due to increased token length. Existing compression approaches—instance-level and token-level—either sacrifice essential local reasoning signals like reflection or yield incoherent outputs. To address these limitations, we propose \methodname, a two-stage chunk-level compression framework that preserves both local information and coherence. Our method segments Long-CoT into manageable chunks, applies LLM-driven inner-chunk compression, and employs an inter-chunk search mechanism to select the short and coherent sequence. Experiments on Qwen2.5-Instruct models across MATH500, AIME24, and GPQA-Diamond demonstrate that \methodname significantly reduces token usage while maintaining comparable reasoning accuracy. On MATH500, \methodname achieves an accuracy of 92.4\%, with only a 0.6\% drop compared to the Long-CoT baseline, while reducing token usage by about 20\%. 
\end{abstract}

\section{Introduction}
Chain-of-Thought (CoT) reasoning \cite{sys1tosys2, mmreasoning,yao2024mulberryempoweringmllmo1like, yao2025r1-ShareVL, yao2025survey} has recently emerged as a powerful technique that enables large language models (LLMs) to perform complex reasoning tasks, such as mathematical problem solving \cite{hendrycksmath2021, adar1} and code generation \cite{llmcode, codesuvery}, by decomposing the reasoning process into a sequence of intermediate steps. Recent advancements, including OpenAI's o1 \cite{o12024}, DeepSeek-R1 \cite{deepseek2024r1lite}, leverage reinforcement learning to scale to Long-CoT, further improving performance and enabling LLMs to tackle real-world tasks. 

However, the extended token length in Long-CoT incurs substantial computational overhead, leading to slower inference and a dramatic increase in KV cache memory usage \cite{sui2025stopoverthinkingsurveyefficient, qu2025surveyefficientreasoninglarge, wang2025harnessingreasoningeconomysurvey}. These factors significantly hinder practical deployment and impose greater demands on hardware infrastructure. Therefore, developing efficient compression methods for Long-CoT that preserve their reasoning capabilities is of critical importance for enabling scalable and deployable reasoning systems.

Existing methods for CoT compression can be broadly categorized into two paradigms: instance-level compression and token-level compression. Instance-level compression includes C3oT \cite{kang2024c3otgeneratingshorterchainofthought} and CoT-Valve \cite{ma2025cotvalvelengthcompressiblechainofthoughttuning}. C3oT utilize powerful LLMs like GPT-4 to directly compress entire CoT sequences. CoT-Valve compresses the length of CoT by identifying and manipulating a specific direction in the parameter space. These methods aim to retain the essential reasoning path while reducing the global token count. 
Token-level compression, such as TokenSkip \cite{xia2025tokenskipcontrollablechainofthoughtcompression}, adopt a more fine-grained strategy by identifying and skipping unimportant tokens. This allows for a compressed representation that retains detailed local information.

However, our evaluation results show that the instance-level compression can degrade the local information by reducing the global token count---the reflection in Long-CoT is reduced, leading to a decline in performance. As reflection is a crucial capability within Long-CoT that enables LLMs to self-reflect and explore the correct answer, it needs to be preserved through a more fine-grained compression approach. TokenSkip as a token-level method could preserve local information such as reflection well by skipping only unimportant tokens. However, through observation and analysis, we find that this direct token-skipping approach often leads to incoherent compressed CoT, creating a gap from the natural language patterns typically used by LLMs.

Based on the above finding, it seems that effectively compressing Long-CoT cannot be achieved solely through instance-level or token-level methods. Therefore, we propose a chunk-level compression approach, which better preserves chunk-level local information and can be implemented via prompting LLMs, thus maintaining linguistic coherence.
However, since each chunk is compressed independently, contextual connections between chunks are lost. A subsequent question is that: 

\begin{quote}
 \vspace{-0.2cm}
 \centering
 \textit{Although coherence within each chunk can be ensured, how to ensure coherence across \textbf{inter chunks}?} 
  \vspace{-0.2cm}
 \end{quote}

Driven by this question, we propose a chunk search mechanism that generates multiple compressed candidate chunks and employs a search model to select the most coherent one. Conditioned on the previously selected optimal chunk, the search model identifies the candidate with the highest likelihood of maintaining continuity, thereby enhancing coherence across the compressed reasoning process. To improve efficiency, we first filter each chunk’s candidates to retain a smaller subset.

\begin{figure*}[t]
\centering
\includegraphics[width=5.5in]{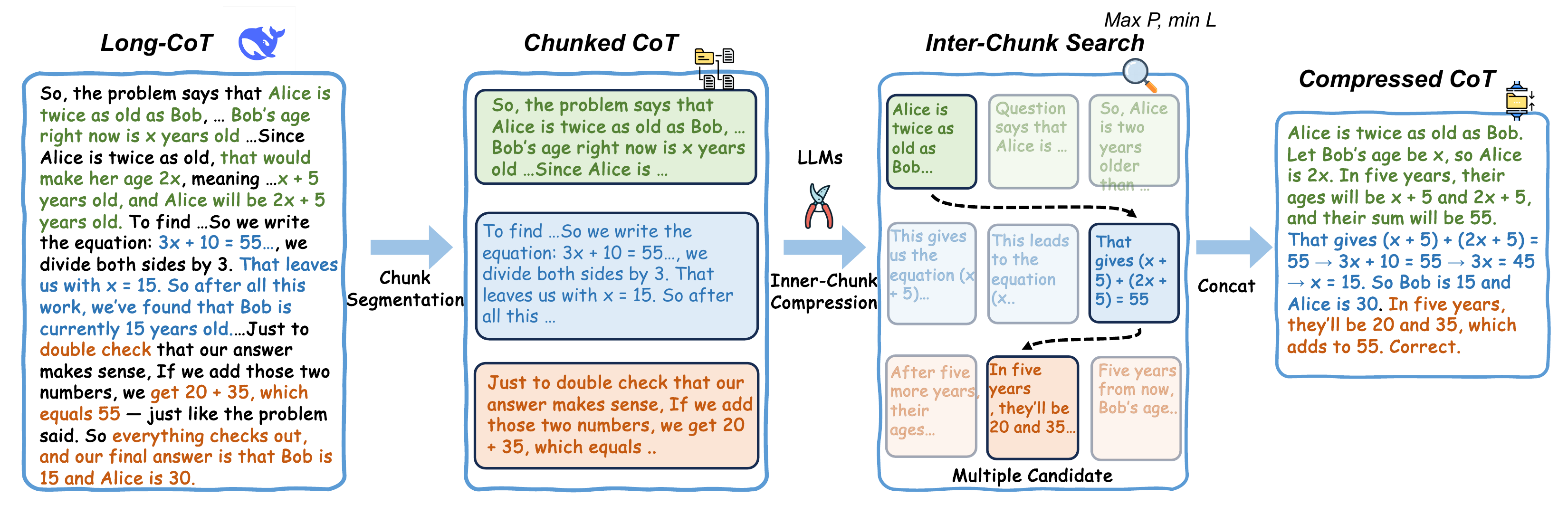}

\caption{Pipeline of our method. The Long-CoT is segmented into chunks, multiple compressed candidates for each chunk are generated using a LLM, and then a compressed CoT is constructed chunk by chunk through inter-chink search with length filtering and probability selection.} 
\label{fig:pipeline}
\vskip -0.2in
\end{figure*}

To this end, we propose \methodname in Figure~\ref{fig:pipeline}, a two-stage method designed to compress Long-CoT on chunk-level: i) The original CoT is segmented into multiple chunks based on predefined length and formatting constraints. Within each chunk, an LLM is prompted to perform local compression. ii) We first generate multiple compressed candidates for each chunk and a chunk-level search is performed to obtain the short and coherent one. By combining inner-chunk compression with inter-chunk search, our method yields a compressed yet consistent CoT, enabling efficient and coherent reasoning. We evaluate our method on Qwen2.5-14B-Instruct and Qwen2.5-32B-Instruct~\cite{qwen2025qwen25technicalreport} on the subset of Open-Math-R1 dataset with responses generated by DeepSeek-R1. Experiments are conducted on the MATH500 \cite{hendrycksmath2021} and AIME24 \cite{AIME2024} benchmarks for mathematical reasoning, and GPQA \cite{rein2023gpqagraduatelevelgoogleproofqa} for out-of-distribution reasoning. Results show that our method consistently reduces inference token usage across model scales and datasets while maintaining comparable accuracy. Our method achieves 92.4\% accuracy on MATH500---only 0.6\% below the Long-CoT baseline (93\%), with about 20\% reduction in token usage (from 2406 to 1949). The contributions of this paper:
\begin{itemize}[leftmargin=*]
   \item 
We find that instance-level compression methods tend to overlook local information---such as reducing the number of reflections in Long-CoT---which negatively impacts performance. In addition, our analysis reveals that token-level methods often lead to CoT lacking coherence.

   \item 
To preserve the local information of Long-CoT and generate coherent reasoning chains, we propose \methodname, a two-stage chunk-level approach. This method combines inner-chunk compression with inter-chunk search to produce CoT that are both efficient and coherent.

   \item 
Extensive results demonstrate that our method can effectively reduce the length of CoT while maintaining the model's reasoning performance across reasoning benchmark.
\end{itemize}

\section{Related Work}

\textbf{Chain-of-Thought.}
\cite{wei2023chainofthoughtpromptingelicitsreasoning,li202512surveyreasoning} prompting has emerged as a powerful technique for improving the reasoning capabilities of large language models (LLMs). By encouraging the model to solve complex problems step by step, CoT significantly enhances the accuracy and interpretability of its outputs. CoT is particularly effective for tasks that requiring multiple solving steps, such as mathematical problem-solving and logical reasoning. Beyond the basic CoT paradigm, many innovative frameworks like Tree of Thought (ToT) \cite{yao2023treethoughtsdeliberateproblem} and Graph of Thought (GoT) \cite{Besta_2024} expand upon the CoT architecture by investigating various reasoning trajectories or integrating structures based on networks. 
Besides, Chain-of-thought reasoning also enable human to comprehend the model's decision-making pathway, thereby rendering the reasoning process both transparent and credible.

\textbf{Efficient Reasoning.}
Some approaches\cite{liu2025efficient} adopt sampling-based and post-training techniques to fine-tune existing Long-CoT models for shorter outputs. For example, Overthinking\cite{overthinking} utilizes DPO and SimPO to construct preference datasets for concise reasoning and trains models accordingly. O1-Pruner\cite{luo2025o1prunerlengthharmonizingfinetuningo1like} establishes baselines for CoT length and accuracy via sampling, then employs offline optimization to shorten CoT without degrading performance. Concise Reasoning \cite{munkhbat2025selftrainingelicitsconcisereasoning} leverages simple fine-tuning strategies based on self-generated concise CoT obtained through best-of-N sampling and few-shot prompting. While other methods use different reasoning paradigm to enhance efficiency. For example, Speculative Thinking\cite{yang2025speculativethinkingenhancingsmallmodel} enables large reasoning models to guide smaller ones during inference at the reasoning level; LightThinker\cite{zhang2025lightthinkerthinkingstepbystepcompression} dynamically compress intermediate thoughts during reasoning and Sleep-time Compute\cite{lin2025sleeptimecomputeinferencescaling} allows models to "think" offline about contexts before queries are presented. Additionally, methods like COCONUT\cite{hao2024traininglargelanguagemodels} and CCOT\cite{cheng2024compressedchainthoughtefficient} enable reasoning in the latent space. Besides, some other work \cite{yang2025dynamicearlyexitreasoning,pan2025learningadaptiveparallelreasoning,ma2025reasoningmodelseffectivethinking,qiao2025conciseconfidenceguidedcompressionstepbystep,zhuang2025acceleratingchainofthoughtreasoninggoalgradient,yang2025thinkneedselfadaptivechainofthought,hou2025thinkprunepruninglongchainofthought,ning2025thoughtsgeneratedequalefficient,li2025tldrlongreweightingefficient,gong2025efficientreasoningchainunconscious} also design novel reasoning paradigms for efficiency.

\textbf{Chain-of-Thought Compression.}
Several methods aim to directly compress Chain-of-Thought (CoT) of Large Reasoning Models. C3oT employs LLMs to compress CoT end-to-end. CoT-Valve\cite{ma2025cotvalvelengthcompressiblechainofthoughttuning} controls the parameter space to generate CoT samples with varying levels of compression for training models that output shorter reasoning paths. TokenSkip\cite{xia2025tokenskipcontrollablechainofthoughtcompression} selectively removes tokens based on their estimated importance within the CoT.

\section{Revisiting Long-CoT Compression}

\subsection{Problem Setup}

\textbf{Long-CoT.} Long-CoT approaches, such as OpenAI's o1 and DeepSeek-R1, exhibit the ability to identify and correct their own mistakes by decomposing complex reasoning steps into simpler subproblems. This iterative process significantly enhances the model’s reasoning capability. Long-CoT typically consists of multiple steps. In this work, we adopt the responses generated by DeepSeek-R1 as representative Long-CoT.
 
\textbf{SFT with Compressed Long-CoT.} We focuses on compressing the token length of Long-CoT by directly reducing the original Long-CoT into shorter reasoning chains. In our setup, the Long-CoT baseline refers to the model fine-tuned using the original Long-CoT responses, while Long-CoT Compression method denotes the model fine-tuned on the compressed versions of Long-CoT. The latter retains the reasoning capabilities of Long-CoT while reducing the number of output tokens.

\subsection{Revisiting Instance-level Compression }

In this section, we investigate existing instance-level compression methods and discuss their limitations in preserving local information, particularly the reflection steps within Long-CoT. Further experimental analysis reveals that this reduction in reflection leads to a decline in performance.

\begin{figure*}[h]
\centering
\includegraphics[width=5.5in]{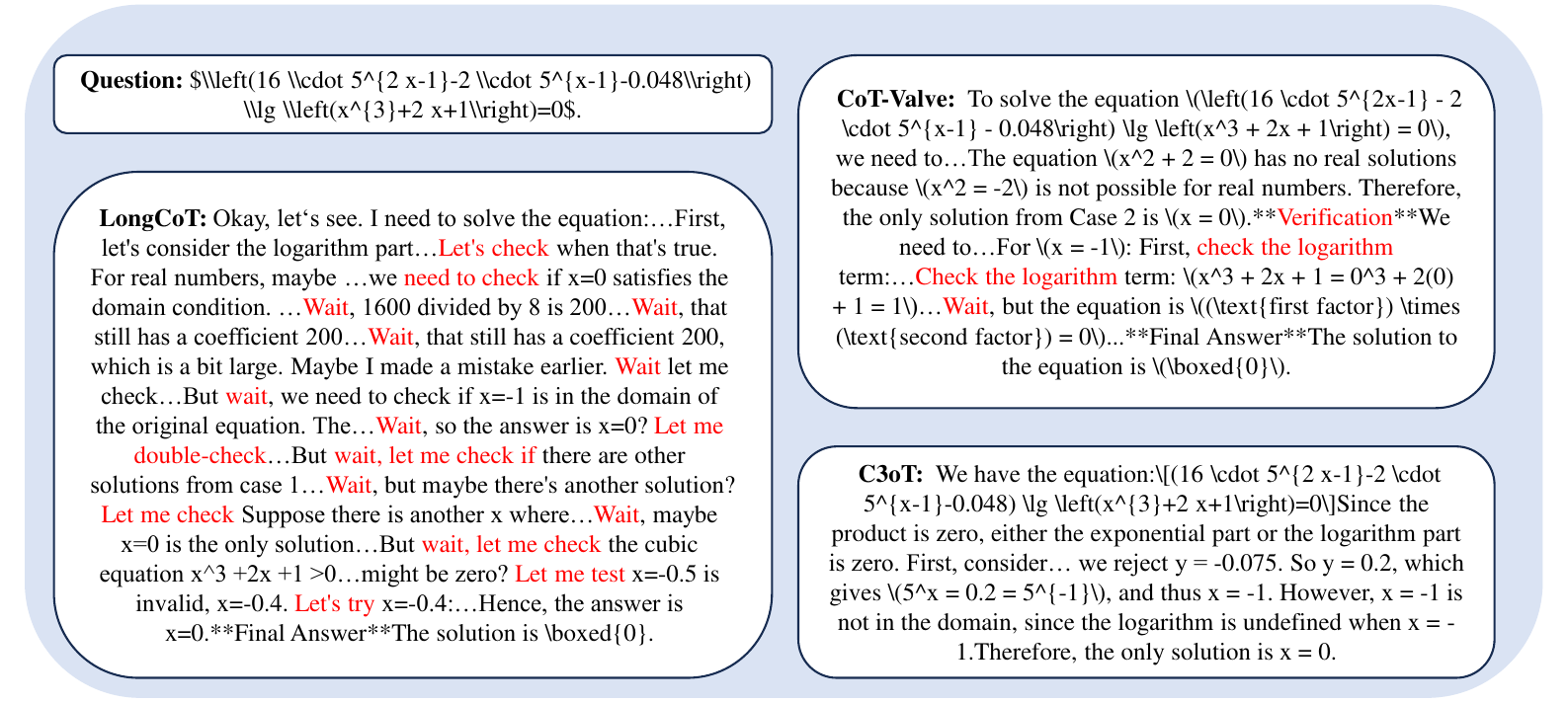}
\caption{Comparison of LongCoT, CoT-Valve, and C3oT. Red text indicates reflection-related phrases such as “Wait”.} 
\label{fig:3_2}
\vskip -0.2in
\end{figure*}

\textbf{Reflection.} 
Reflection is the model’s ability to evaluate and revise its reasoning process during problem-solving. It enables the model to recognize and correct its own mistakes, decompose complex steps into simpler components, and adapt its strategy when the current approach proves ineffective. This iterative process of self-assessment and adjustment plays a crucial role in enhancing the model’s overall reasoning capability.

We select the C3oT that simply prompting the LLMs to obtain the compressed CoT, CoT-Valve that manipulating a specific direction in the parameter space to reduce the length of CoT to Compress the Long-CoT from DeepSeek-R1. As shown in Figure~\ref{fig:3_2}, we find that the compressed CoT obtained through these two methods are able to preserve certain key steps and reach the final answer. However, compared to the original Long-CoT, they omit a considerable number of intermediate steps and exploratory attempts—particularly the processes of reflection and strategy switching that are often essential for arriving at the correct solution. Since reflection is a critical reasoning skill that models are expected to learn from Long-CoT supervision, the absence of this capability prompts an important question:

\begin{quote}
 \vspace{-0.2cm}
 \centering
 \textit{Does the reduced frequency of reflection in compressed Long-CoT adversely affect the performance of models fine-tuned on it?}
  \vspace{-0.2cm}
 \end{quote}

To evaluate this, we count the occurrences of indicative reflection-related keywords (See Sec.~\ref{sec:reflection exp} for more details,) that is also adopted by other methods~\cite{chen2025seal}. We calculate the average number of reflection steps in 500 examples for both the original Long-CoT and the CoT compressed by the two mentioned methods. We then evaluate the performance of models fine-tuned using these different CoT. As shown in Table~\ref{tab:reflection_fingding}, the quantitative results reveal that as the number of reflection steps decreases, the performance of the fine-tuned model deteriorates—indicating that the absence of reflection impairs the model’s reasoning ability.


\begin{wraptable}{r}{0.55\textwidth}
\vspace{-0.5cm}
\caption{Comparison of methods on average reflection and accuracy on MATH500.}
\label{tab:reflection_fingding}
\begin{tabular}{lcc}
\toprule
\textbf{Method} & \textbf{Avg. Reflection} & \textbf{Accuracy (\%)} \\
\midrule
Long-CoT & 18.68 & 88.0 \\
C3oT & 0.15 & 65.8 \\
CoT-Valve & 8.36 & 77.4 \\
\bottomrule
\end{tabular}
\vskip -0.2in
\end{wraptable}

\textbf{Conclusion.} Instance-level compression methods operate from a global perspective and fail to preserve local information such as reflection, which is crucial for reasoning. Since the presence of reflection significantly affects the reasoning ability of fine-tuned models, a more fine-grained compression strategy is needed to effectively retain such local information.

\subsection{Revisiting Token-level Compression}

Token-level methods are inherently capable of preserving local information in Long-CoT, such as reflection. In this section, we explore existing token-level compression approaches and analyze the coherence issues observed in the resulting CoT. Furthermore, we perform a quantitative analysis using loss values derived from probabilistic predictions.
\begin{figure*}[h]
\centering
\includegraphics[width=5.5in]{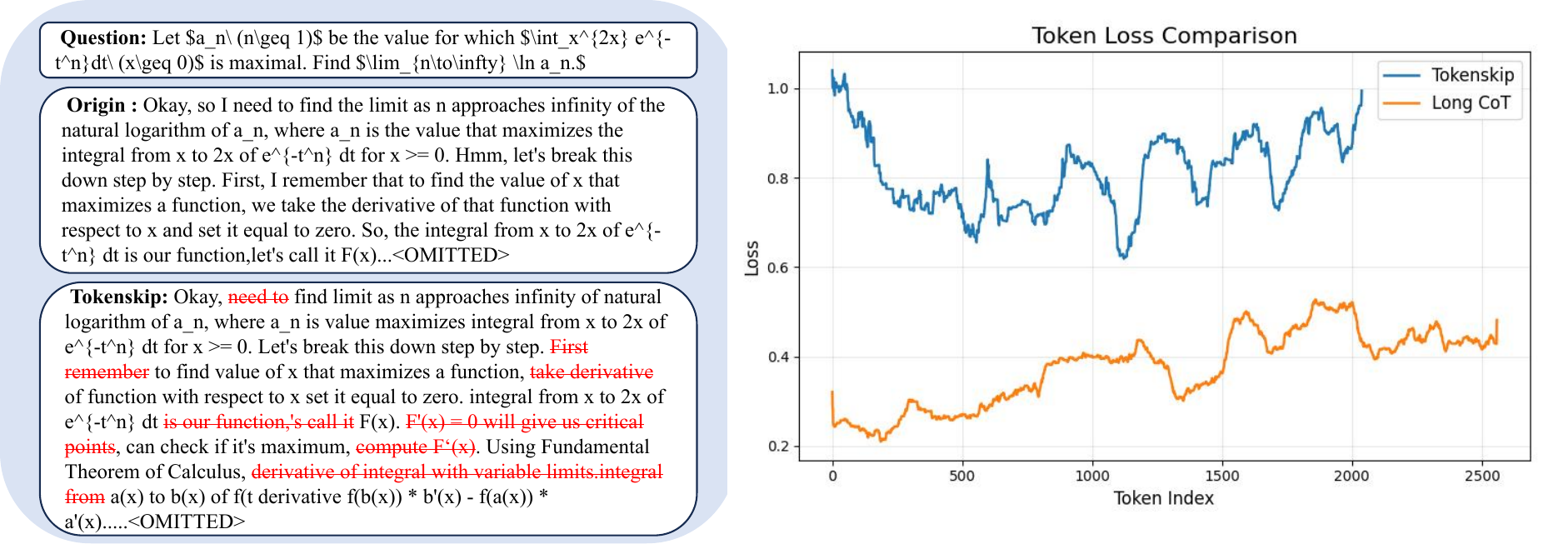}
\caption{Left: Example of TokenSkip CoT Compression, Right: Token-level loss curves of Long-CoT and TokenSkip.} 
\label{fig:3_3}
\vskip -0.2in
\end{figure*}

We select TokenSkip as a representative token-level method to compress Long-CoT generated by DeepSeek-R1. As shown in Figure~\ref{fig:3_3} Left, although TokenSkip can identify and remove unimportant tokens---thus partially preserving the original semantic content---we observe that the compressed CoT often exhibit clear incoherence, for example, "\texttt{ of function}" and "\texttt{, can check if}". We attribute this to a mismatch between the compressed outputs and the natural language patterns expected by LLMs. This gap not only results in incoherent outputs after supervised fine-tuning, but also affects the training dynamics by increasing the prediction loss due to the unnatural text of the input.

We quantify coherence using token-level loss, computed as the negative log-likelihood of each token in the compressed CoT predicted by the LLM. (See more details in Sec.~\ref{sec:coherence exp}). As shown in Figure~\ref{fig:3_3} Right, the token-level loss of the Tokenskip is generally higher than that of the origin, which further indicates a significant inconsistency between its output and the original content, making it less aligned with the typical output patterns of LLMs.

\textbf{Conclusion.} Token-level compression methods often produce incoherent CoT, which can negatively impact the training process and lead to models that generate incoherent outputs. In contrast, instance-level methods, such as C3oT, compress global information through prompting LLMs, resulting in more coherent outputs.

\subsection{Derived Insight}

Based on the above analysis, we argue that effectively compressing Long-CoT cannot be achieved solely through instance-level or token-level methods. To address this, we propose compressing Long-CoT at the \textbf{chunk level}. Chunk-level compression allows for better preservation of local information within each chunk while maintaining stronger inner-chunk coherence. To ensure coherence across chunks, we introduce a inter-chunk search mechanism, which selects the most coherent sequence of chunks. Additionally, we incorporate a search over compression lengths to further enhance compression efficiency.

\section{Method}

\subsection{Chunk Segmentation}
We are given a dataset of problem-solution pairs, denoted as $\mathcal{D} = \{(x_k, y_k)\}_{k=1}^N$
where \(x_k\) represents a problem and $y_k = [y_k^1, \ldots, y_k^{m_k}]$ denotes its corresponding solution generated by a large language model (LLM) parameterized by \(\boldsymbol{\theta}\), denoted as \(\pi_{\boldsymbol{\theta}}(\cdot \mid x_k)\). Each solution \(y_k\) (CoT) is segmented into a sequence of \(m_k\) constituent chunks: $y_k = [c_{k,1}, c_{k,2}, \ldots, c_{k,m_k}]$

To obtain the chunks \(c_{k,j}\) from the raw text \(y_k\), we use the following segmentation strategy:\textbf{Minimum length requirement}: A chunk must contain at least a predefined minimum number of characters or tokens (e.g., 50 words).
\textbf{Double newline boundary}: A chunk ends when two consecutive newline characters (`/n/n`) are encountered, provided that the current chunk has met the minimum length requirement. This results in variable-length chunks that are semantically meaningful and structurally coherent, often corresponding to paragraphs or logical substeps in a solution. This chunking strategy ensures that each \(c_{k,j}\) captures a complete unit of reasoning or explanation, which is essential for later compression and search.

\subsection{Inner-Chunk Compression}
The simplification process for a given pair $(x, y)$ (dropping the index $k$ for clarity) involves the following steps: For each chunk $c_i$ in the solution $y = [c_1, c_2, \ldots, c_m]$, we utilize a separate LLM, parameterized by $\mathbf{\phi}$ and denoted as $\pi_{\mathbf{\phi}}$, to generate multiple simplified candidate versions. Given the original chunk $c_i$ and a suitable prompt $p$, we sample $M$ candidate simplified chunks from the conditional distribution $\pi_{\mathbf{\phi}}(\cdot | p, c_i)$. These candidates for chunk $c_i$ are denoted as $\{ \hat{c}_i^j \}_{j=1}^M$.
\begin{equation}
\hat{c}_i^j \sim \pi_{\mathbf{\phi}}(\cdot \mid p, c_i), \quad \text{for } j=1, \ldots, M
\end{equation}
This process is applied independently to each chunk $c_i$ of the original solution $y$. The prompt \(p\) is carefully designed to guide the LLM toward generating simplified and concise versions of the input chunk while preserving its original meaning. The full prompt used in our experiments is provided in the Appendix.

\subsection{Inter-Chunk Search}
After obtaining $M$ candidate simplified chunks $\{ \hat{c}_i^j \}_{j=1}^M$ for each original chunk $c_i$ in $y$, we aim to construct a complete simplified solution sequence $y^* = [\hat{c}_1^*, \hat{c}_2^*, \ldots, \hat{c}_m^*]$ by selecting one optimal candidate $\hat{c}_i^*$ for each position $i$. The selection criteria prioritize brevity and a low "loss", where loss is inversely related to the probability assigned by the original LLM $\pi_{\mathbf{\theta}}$ to the simplified sequence. We employ a greedy search approach:

\textbf{Length-based Filtering:} For each position $i$, we first filter the set of $M$ candidates $\{ \hat{c}_i^j \}_{j=1}^M$. We discard the $\alpha \cdot M$ longest candidates, keeping the $(1-\alpha)M$ shortest ones, where $\alpha \in [0, 1)$ is a predetermined filtering ratio. Let the filtered set of candidates for position $i$ be $\tilde{\mathcal{C}}_i \subseteq \{ \hat{c}_i^j \}_{j=1}^M$.

\textbf{Probability-based Selection:} We iteratively select the best simplified chunk for each position $i=1, \ldots, m$. At position $i$, having selected the optimal simplified chunks $\hat{c}_1^*, \ldots, \hat{c}_{i-1}^*$ for the preceding positions, we choose the candidate $\hat{c}_i^* \in \tilde{\mathcal{C}}_i$ that maximizes the conditional probability under the original LLM $\pi_{\mathbf{\theta}}$, given the original problem $x$ and the previously selected simplified chunks:
\begin{equation}
\hat{c}_i^* = \arg\max_{\hat{c} \in \tilde{\mathcal{C}}_i} \pi_{\mathbf{\theta}}(\hat{c} \mid x, \hat{c}_1^*, \ldots, \hat{c}_{i-1}^*)
\label{eq:selection_i}
\end{equation}

For the first chunk ($i=1$), the selection is based solely on the probability conditioned on the problem $x$:

\begin{equation}
\hat{c}_1^* = \arg\max_{\hat{c} \in \tilde{\mathcal{C}}_1} \pi_{\mathbf{\theta}}(\hat{c} \mid x)
\label{eq:selection_1}
\end{equation}

\subsection{Compressed CoT}
The final simplified solution $y^*$ for the problem $x$ is constructed by concatenating the sequence of optimally selected simplified chunks:
\begin{equation}
y^* = [\hat{c}_1^*, \hat{c}_2^*, \ldots, \hat{c}_m^*]
\label{eq:final_output}
\end{equation}

This entire process is applied to each $(x_k, y_k)$ pair in the dataset $\mathcal{D}$ to obtain a dataset $\mathcal{D}_{\text{compressed}}$ of simplified solutions.

\subsection{Fine-tuning with Compressed CoT}
After obtaining the compressed dataset $\mathcal{D}_{\text{compressed}} = \{(x_k, y_k^*)\}_{k=1}^N$, we perform standard supervised fine-tuning (SFT) on the base model $\pi_{\theta}$ to better align it with the simplified reasoning trajectories. The training objective is to maximize the log-likelihood of the compressed outputs given the input problems:
\begin{align}
    \mathcal{L}_{\text{SFT}}(\theta) = \sum_{k=1}^N \log \pi_{\theta}(y_k^* | x_k)
\end{align}
This fine-tuning step encourages the model to generate concise yet faithful reasoning chains.

\section{Experiments}
\subsection{Experiment Settings}

\textbf{Dataset.} For training, we use the OpenR1-Math-220k dataset, a large-scale benchmark for mathematical reasoning. It consists of 220k math problems, each responses generated by DeepSeek-R1. For evaluation, we leverage two widely used mathematical reasoning benchmarks. MATH500 and AIME24. GPQA-Diamond as an out-of-distribution benchmark. More details are in Appendix~\ref{apd:benchmark}.

\textbf{Baseline.} We consider two primary baselines: \textbf{CoT-Valve} and \textbf{TokenSkip}.  
\textbf{\methodname$_{random}$} is a variant of \methodname that randomly selects a candidate chunk during compression.  
\textbf{Long-CoT} refers to supervised fine-tuning (SFT) on the original DeepSeek-R1 responses without any compression.  
\textbf{Base} denotes the model without SFT. In our experiments, the base model is fine-tuned on the compressed CoT generated by each method. More details can be found in Appendix~\ref{apd:baseline}.

\textbf{Metric.} We employ the following three metrics to evaluate the model’s performance.
\textbf{Accuracy}: For MATH500 and GPQA-Diamond, we report pass@1 accuracy. For AIME24, due to its small size, we report avg@10 accuracy.
\textbf{Token} (Token Length): The average token length of generated responses.
\textbf{Valid} (Valid Token Length): The average token length of responses that are answered correctly.

\textbf{Implementation Details.} We primarily evaluate our method using the Qwen2.5-Instruct series (14B/32B). All evaluations are conducted using the \texttt{lighteval} framework, following the widely adopted Long-CoT evaluation setting, with a temperature of 0.6 and a top-p of 0.95. 

For supervised fine-tuning (SFT), we use a learning rate of 1e-5 and train for 4 epochs using the \texttt{LlamaFactory} library. For chunk compression, we utilize LLaMA3.1-70B-Instruct with a sampling temperature of 0.75 and generate 8 candidate chunks. The chunk search method is performed by the DeepSeek-R1-Distill-Qwen-14B model, More details are in Appendix~\ref{apd:details}.

\begin{table}[h]
    \centering
    \renewcommand{\arraystretch}{1.6}
    \Huge
    \caption{Main experiment results. We present the performance of two models and report accuracy ($\uparrow$), average token length (Token) ($\downarrow$), valid token length (Valid) ($\downarrow$) on three reasoning benchmark.}
    \vskip 0.1in
    \resizebox{\textwidth}{!}{%
        \begin{NiceTabular}{l cc cc cc}
        \toprule
        \textbf{Methods}
          & \multicolumn{2}{c}{\textbf{MATH500}}
          & \multicolumn{2}{c}{\textbf{AIME24}}
          & \multicolumn{2}{c}{\textbf{GPQA-Diamond}} \\
        \cmidrule(lr){2-3}\cmidrule(lr){4-5}\cmidrule(lr){6-7}
          & Accuracy
          & Token (Valid) 
          & Accuracy 
          & Token (Valid) 
          & Accuracy 
          & Token (Valid)  \\
        \midrule
        {\textit{\textbf{Qwen2.5-14B-Instruct}}} & & & & & & \\
        Base          & 79.8 & -       & 11.00 & -       & 47.97 & -      \\
        Long-CoT      & 88.0 & 3781 (2601) & 30.00 & 12101(6402) & 51.51 & 9600 (7830) \\
        \midrule
        CoT-Valve     & 77.4 & 3733 (\textbf{1343}) & 15.00 & 12972 (\textbf{4186}) & 39.89 & 10257 (\textbf{6704}) \\
        TokenSkip     & 82.8 & 4236 (2313)          & 17.66 & 13504 (4644)       & 33.83 & 11974 (8130) \\
        \midrule
        \methodname$_{random}$ & 81.2 & 3880 (2033) & 24.00 & 12444 (6381) & 48.48 & 9524 (7153) \\
        \rowcolor{Highlight}
        \methodname   & \textbf{84.8} & \textbf{3369} (2074)
                      & \textbf{25.66} & \textbf{11369} (5575)
                      & \textbf{49.49} & \textbf{8544} (6962) \\
        \midrule
        \midrule
        {\textit{\textbf{Qwen2.5-32B-Instruct}}} & & & & & & \\
        Base          & 83.2 & -       & 16.66 & -       & 50.0  & -      \\
        Long-CoT      & 93.0 & 3147 (2406) & 50.66 & \textbf{10541} (5997) & 61.11 & 8054 (6199) \\
        \midrule
        CoT-Valve     & 91.0 & 2718 (1891) & 39.33 & 11357 (5898) & 54.04 & 9578 (6891) \\
        TokenSkip     & 89.8 & 3004 (\textbf{1871}) & \textbf{44.33} & 10881 (6000) & \textbf{59.59} & 8505 (5877) \\
        \midrule
        \methodname$_{random}$ & 89.8 & 2899 (1965) & 42.00 & 11135 (5705) & 54.04 & 8335 (6510) \\
        \rowcolor{Highlight}
        \methodname   & \textbf{92.4} & \textbf{2661} (1949)
                      & 43.33 & 10747 (\textbf{5495})
                      & 59.09 & \textbf{6963} (\textbf{5005}) \\
        \bottomrule
    \end{NiceTabular}%
    }
    \label{tab:main}
    \vskip -0.1in
\end{table}
\subsection{Main Results}

For the Qwen2.5-14B-Instruct model in Table~\ref{tab:main}, we observe that \methodname achieves consistent improvements over the Long-CoT baseline by significantly reducing the average token length—ranging from a reduction of 412 tokens on MATH500 to 1056 tokens on GPQA-Diamond. Importantly, \methodname attains the highest accuracy and lowest token length across all three benchmarks comparing with other baselines, demonstrating its ability to effectively compress Long-CoT without compromising its reasoning effectiveness. Our method also performs well on the out-of-distribution benchmark GPQA-Diamond, highlighting its generalizability. Compared to \methodname$_{random}$, our full method further improves both accuracy and token efficiency, validating the effectiveness of the proposed inter-chunk search in selecting shorter and more coherent CoT.

As we scale up to the Qwen2.5-32B-Instruct model, \methodname continues to outperform all baselines in terms of token length while achieving the best or comparable accuracy. On MATH500, our method achieves a strong accuracy of 92.4\%, with only 0.6\% performance drop compared to the Long-CoT baseline (93.0\%), while reducing the average token length by around 500 tokens. Furthermore, the valid token length is reduced by approximately 20\% (from 2406 to 1949) under nearly equal numbers of correct responses. The consistent improvement over \methodname$_{random}$ on the larger model further supports the robustness and scalability of our proposed search strategy.

\subsection{Reflection Evaluation}
\label{sec:reflection exp}

We conduct this analysis by counting the occurrences of reflection-related keywords: “wait”, “alternatively”, "emm", "hmm". These tokens indicate shifts in reasoning or self-reflection.

As shown in Table~\ref{tab:reflection_ours}, our method preserves significantly more reflection steps compared to other baselines---on average, six more than CoT-Valve. We preserve approximately 78\% of the reflections found in Long-CoT, while achieving accuracy competitive with the original Long-CoT responses.

\begin{table}[h]
\centering
\begin{minipage}[t]{0.48\textwidth}
\vspace{-0.5cm}
\caption{Comparison of methods on average reflection and accuracy on MATH500.}
\vskip 0.1in
\label{tab:reflection_ours}
\resizebox{\linewidth}{!}{%
\begin{tabular}{lcc}
\toprule
\textbf{Method} & \textbf{Avg. Reflection} & \textbf{Accuracy (\%)} \\
\midrule
Long-CoT & 18.68 & 88.0 \\
C3oT & 0.15 & 65.8 \\
CoT-Valve & 8.36 & 77.4 \\
\rowcolor{Highlight}
\methodname & \textbf{14.59} & \textbf{84.8} \\
\bottomrule
\end{tabular}
}
\vskip 0.2in
\end{minipage}%
\hfill
\begin{minipage}[t]{0.45\textwidth}
\vskip -0.2in
\caption{Coherence evaluation. Comparison of methods on token-level loss.}
\vskip 0.1in
\label{tab:token_loss}
\resizebox{\linewidth}{!}{%
\begin{tabular}{lc}
\toprule
\textbf{Method} & \textbf{Token-Level Loss} \\
\midrule
Long-CoT & 0.41 \\
TokenSkip & 0.87 \\
\methodname$_{random}$ & 0.63 \\
\rowcolor{Highlight}
\methodname & \textbf{0.59} \\
\bottomrule
\end{tabular}
}
\vskip -0.6in
\end{minipage}
\end{table}

\subsection{Coherence Evaluation}
\label{sec:coherence exp}

To quantitatively assess the coherence of compressed CoT, we compute the token-level log-likelihood loss using the DeepSeek-R1-Distill-Qwen-14B model. Specifically, given a compressed CoT as input and the original uncompressed CoT as reference, we evaluate the average token-level loss for TokenSkip, \methodname$_{random}$, \methodname.

Table~\ref{tab:token_loss} reports the average token-level loss for each method. The results indicate that both of our methods achieve lower token-level loss compared to TokenSkip, indicating better coherence between tokens in the compressed CoT. Moreover, \methodname achieves lower loss than \methodname$_{random}$, demonstrating that the introduction of the search mechanism improves inter-chunk coherence. Enhanced coherence contributes to greater stability during training and enables the fine-tuned model to produce more semantically precise outputs.

\begin{figure}[h]
\vskip -0.1in
    \centering
    \begin{subfigure}{0.495\linewidth}
        \includegraphics[width=\linewidth]{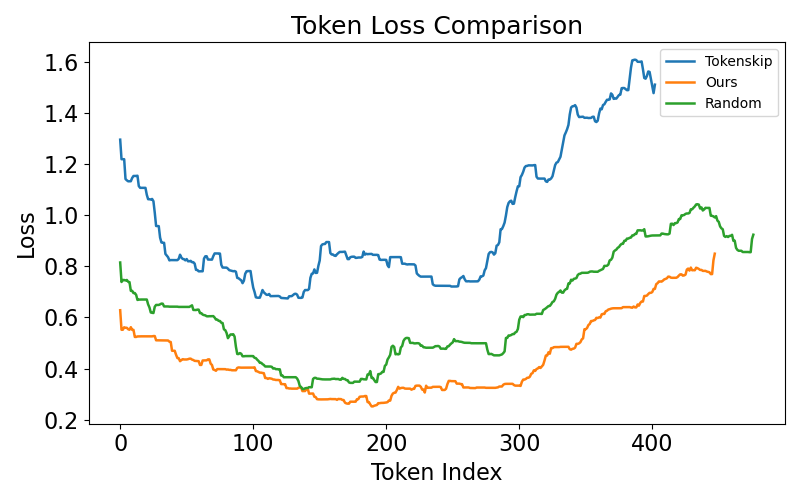}
        \caption{Token loss comparison on case 1}
    \end{subfigure}
    \hfill
    \begin{subfigure}{0.495\linewidth}
        \includegraphics[width=\linewidth]{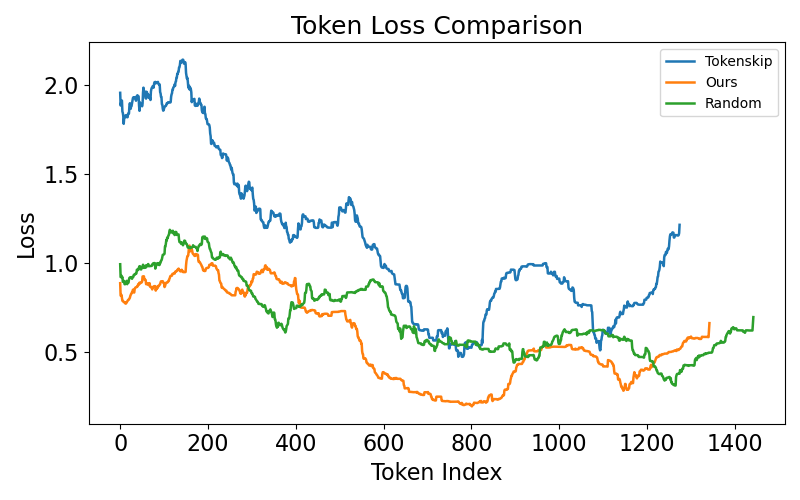}
        \caption{Token loss comparison on case 2}
    \end{subfigure}
    \caption{Token-level loss visualization.}
\vskip -0.2in
\label{fig:loss_tokenskip}
\end{figure}

To complement the quantitative results, we also perform a qualitative analysis by visualizing token-level loss across several representative examples in Figure~\ref{fig:loss_tokenskip}. We observe that TokenSkip frequently exhibits high-loss regions, particularly around intermediate conclusions and reflective reasoning steps. \methodname$_{random}$ and \methodname achieves lower loss compared to \methodname$_{random}$ by leveraging the search mechanism to identify chunks that are more coherent within the given context.

\subsection{Ablation Study}

\begin{wraptable}{r}{0.5\textwidth}
\caption{Ablation study on chunk size.}
\begin{tabular}{c c c }
\toprule
\textbf{Chunk Size} & \textbf{MATH500} & \textbf{AIME24}\\
\midrule
1000 & 79.0 (4188) & 21.66 (13074)\\
500 & \textbf{81.2} (\textbf{3880}) & \textbf{24.00} (\textbf{11369})\\
\bottomrule
\end{tabular}
\label{tab:ab chunksize}
\vskip -0.1in
\end{wraptable}

\textbf{Chunk Size.} Table~\ref{tab:ab chunksize} reports the results of using different chunk size constraints during chunk segmentation. For a clearer ablation, we compare variants without the search mechanism, i.e., \methodname$_{random}$. The results show that smaller chunk sizes yield higher-quality compressed CoT, as finer-grained chunks better preserve local information and reduce the compression difficulty for LLMs. In the limit where the chunk size becomes unbounded, the method effectively reduces to C3oT.

\begin{wraptable}{r}{0.55\textwidth}
\vskip -0.1in
    \caption{Ablation study on search model. 
    ``w/o'' denotes the absence of a search model, 
    ``w/ Qwen'' uses Qwen2.5-14B-Instruct as the search model, 
    and ``w/ DeepSeek-Distill'' uses DeepSeek-R1-Distill-Qwen-14B as the search model.}
    \label{tab:ab search}
    \begin{tabular}{lcc}
        \toprule
        \textbf{Methods} 
        & \multicolumn{2}{c}{\textbf{MATH500}} \\
        & Accuracy  & Token(Valid) \\
        \midrule
        \textbf{Qwen2.5-14B-Ins} & &  \\
        w/o & 81.2 & 3880 (2033) \\
        w/ Qwen & 83.0 & 3373 (\textbf{1874}) \\
        w/ DeepSeek-Distill & \textbf{84.8} & \textbf{3369} (2074) \\
        \bottomrule
    \end{tabular}%
    \vskip -0.1in
\end{wraptable}

\textbf{Search Model.} We investigate the impact of different models used in the search phase of reasoning compression. Specifically, we compare Qwen2.5-14B-Instruct (Qwen) and DeepSeek-R1-Distill-Qwen-14B (DeepSeek-Distill) as the search model. As shown in Table~\ref{tab:ab search},  we observe that both models, when used as the search model, improve accuracy compared to the variant without search. Specifically, DeepSeek-Distill tends to favor longer responses, resulting in a larger gain in accuracy, while Qwen prefers shorter responses, leading to a lower valid token length. Overall, both search models contribute to improved performance, and by selecting shorter yet coherent chunks, the search process ultimately leads to reduced total token usage.

\vskip -0.4in
\section{Conclusion}
\vskip -0.1in
In this paper, we propose \methodname, an effective framework for compressing long Chain-of-Thought (CoT) reasoning by combining inner-chunk compression with an inter-chunk search mechanism. Unlike existing approaches that either compromise critical reasoning behaviors—such as reflection—or lead to incoherent outputs, \methodname effectively reduces token length while maintaining high reasoning quality. Experimental results across multiple benchmarks demonstrate the method's ability to preserve performance under significant compression. These findings underscore the potential of chunk-level CoT compression as a practical and scalable solution for enhancing the efficiency and deployability of large-scale reasoning models.


\newpage
\bibliography{reasoning}
\bibliographystyle{abbrv}


\appendix

\newpage
\section{More Details about Experiments}

\subsection{Implementation Details.} 
\label{apd:details}
We perform compression using LLaMA3.1-70B-Instruct on 4 × 80GB GPUs. For model training, we conduct full-parameter fine-tuning of Qwen2.5-14B-Instruct on 8 × 80GB GPUs and Qwen2.5-32B-Instruct on 16 × 80GB GPUs. All fine-tuning procedures run for more than 2–4 hours. The hyperparameters used for full fine-tuning are summarized in Table~\ref{tab:hyperparameters}.

\begin{table}[ht]
\centering
\caption{Hyperparameters for the Qwen2.5-14B-Instruct and Qwen2.5-32B-Instruct.}
\vskip 0.1in
\label{tab:hyperparameters}
\begin{tabular}{lcc}
\hline
\textbf{Hyperparameter}   & \textbf{Qwen2.5-14B-Instruct} & \textbf{Qwen2.5-32B-Instruct} \\ \hline
cutoff\_len               & 8192                 & 8192                      \\ 
batch\_size               & 8                   & 2                        \\ 
learning\_rate            & 1.0e-5               & 1.0e-5                    \\ 
num\_train\_epochs        & 4.0                  & 4.0                       \\ 
lr\_scheduler\_type       & cosine               & cosine                    \\ 
warmup\_ratio & 0.1 & 0.1 \\
\hline
\end{tabular}
\end{table}

\subsection{Benchmark.}
\label{apd:benchmark}

\textbf{MATH500}: A challenging math dataset comprising 500 problems from high school math competitions.

\textbf{AIME24}: A benchmark dataset consisting of 30 challenging mathematical problems from the 2024 American Invitational Mathematics Examination. 

\textbf{GPQA-Diamond}: A high-difficulty subset of the GPQA benchmark, with 198 complex graduate-level multiple-choice questions across various scientific domains.

\subsection{Baseline.}
\label{apd:baseline}

\textbf{CoT-Valve}: We adopt the Short-Long-Short CoT compression strategy proposed by CoT-Valve, which aligns with our experimental setting. We use an untrained model as the short model and a model fine-tuned on Long-CoT as the long model. By applying model merging, we obtain a Short-Long-Short model, following the setup introduced in CoT-Valve. Specifically, we perform linear interpolation with weights of (0.9, 0.1) and (0.8, 0.2) for the short and long models, respectively, to create different variants of the Short-Long-Short model. These merged models are then used to sample and construct the MixChain of Short-Long-Short CoT dataset.

\textbf{TokenSkip}: This baseline directly applies token-level compression to Long-CoT to generate shortened CoT. We follow its setting to measure the  token importance by LLMLingua-2 compressor. The control ratios that we use are 0.9, 0.8, 0.7, 0.6.

\textbf{C3oT}: In Table~\ref{tab:reflection_fingding}, we adopt the prompt template provided in the original implementation and use the same LLM (LLaMA3.1-70B-Instruct) as the compressor to ensure a fair comparison.

\subsection{Metric.}

\textbf{Accuracy.} For MATH500 and GPQA-Diamond, we report \textbf{pass@1} accuracy, where the model is evaluated with a single response. For AIME24, due to its small size (30), we report the accuracy \textbf{avg@10}, calculated as the average accuracy over 10 independent runs.

\textbf{Token (Token Length)}: The average token length of all model-generated responses, used to evaluate the overall compression effectiveness.

\textbf{Valid (Valid Token Length)}: The average token length of responses that are answered correctly. This metric is introduced to better analyze the relationship between output length and successful reasoning.

\subsection{Training Dataset.}
 
We use the OpenR1-Math-220k dataset, a large-scale benchmark for mathematical reasoning. It consists of 220k math problems, each accompanied by two to four reasoning traces generated by DeepSeek R1 for problems sourced from NuminaMath 1.5. All traces have been verified using Math Verify. We randomly sample 5,000 examples from it. 

\subsection{Filter Strategy.}
\label{apd:filter}
Each response from DeepSeek-R1 is first segmented into multiple chunks using our chunk segmentation strategy. To ensure efficient downstream compression, we filter out samples with more than 30 chunks, reducing the initial 5k samples to 3.8k. We further refine the dataset by verifying \texttt{has\_vaild\_answer}: Whether the original R1 response contains an extractable answer and \texttt{has\_same\_answer}: Whether the answer extracted from the compressed CoT matches the original one. Additionally, we remove samples with excessively low or high compression ratios. After this filtering process, a total of 2,513 samples are retained for training.

\section{More Discussion}

\subsection{Necessity of Chunk.}

C3oT compresses Long-CoT directly via LLMs. However, due to the extremely long context of Long-CoT, LLMs often struggle to follow instructions faithfully and preserve critical information. Specifically, we use the advanced model LLaMA3.1-70B-Instruct as the compressor. As shown in Table~\ref{tab:c3ot_filter}, after compressing 3,620 Long-CoT samples, only 442 resulting CoT retain answers consistent with the original responses. This outcome highlights the limitations of direct instance-level compression and underscores the necessity of our proposed chunk-level approach.

\begin{table}[h]
\centering
\caption{Filtering statistics of C3oT-compressed data based on answer consistency. \texttt{has\_same\_answer} is introduce in Appendix~\ref{apd:filter}.}
\vskip 0.1in
\label{tab:c3ot_filter}
\begin{tabular}{lcc}
\toprule
\textbf{Stage} & \textbf{Sample Count} \\
\midrule
Before \texttt{has\_same\_answer} filter & 3,620 \\
After \texttt{has\_same\_answer} filter  & 442 \\
\bottomrule
\end{tabular}
\end{table}


\subsection{Main Results.}
As shown in Table~\ref{tab:main}, our method achieves a substantial reduction of over 1,000 tokens on the GPQA-Diamond benchmark, with only a minimal performance drop (approximately 2\%) compared to the Long-CoT baseline. The strong performance on this out-of-distribution benchmark suggests that models may inherit overly verbose reasoning patterns from Long-CoT supervision, which are then reflected in other tasks. This observation highlights the practical significance of compressing Long-CoT.

Additionally, we observe that on AIME24, the 32B model exhibits a noticeable reduction in valid token length, while the overall token length remains nearly unchanged. This is because AIME24 is a highly challenging task, and when the model produces incorrect answers, it tends to generate longer responses. Thus, although valid reasoning becomes more concise, the total output length does not decrease accordingly.

\section{Prompt Template}

The compression prompt used for LLMs is provided in Table~\ref{tab:prompt compress}. The templates for dataset construction, mathematical evaluation and GPQA evaluation are shown in Table~\ref{tab:mathtemplate}.


\begin{table}[h]
  \centering
  \caption{Compression prompt for LLMs.}
\vskip 0.05in
  \begin{tabular}{p{13cm}}
    \hline
    Here is an reasoning piece excerpt from some math problem solving process (it is incomplete, but this doesn't matter.): \{step\}

    \textbf{Instructions:}

    You need to simplify the wording of given reasoning piece to get a concise reasoning piece.

    \textbf{Notice:}\\
    1. Avoid omitting any reasoning steps. You should keep all the reflection, analysing, checking steps and even steps making mistakes. (Especially steps contains word “wait”, “hmm”)\\
    2. Directly give me the simplified content without any additional words.\\
    3. Do not add additional steps or continue the reasoning process.\\
    4. Follow the format of given reasoning piece.

    \textbf{Output format:}
    \texttt{<start>}
    (simplified content)
    \texttt{<end>} \\ 
    \hline
  \end{tabular}
  \label{tab:prompt compress}
\end{table}


\begin{table}[h]
  \centering
\caption{Template for Dataset construction and Evaluation.}
\vskip 0.05in
  \begin{tabular}{p{13cm}}
    \hline
    \textbf{Dataset construction template:}\\
    \{Question\} Let's think step by step and output the final answer within boxed\{\{\}\}, \{Compressed\_CoT\}.\\
        
    \textbf{MATH template:}\\
    \{Question\} Let's think step by step and output the final answer within boxed\{\{\}\}.\\

    \textbf{GPQA template:}\\
    \{Question\} Answer the following multiple choice question. The last line of your response should be of the following format: 'Answer: \$LETTER' (without quotes) where LETTER is one of ABCD. Let's think step by step before answering.\\
A) {A}\\
B) {B}\\
C) {C}\\
D) {D}\\
    \hline
  \end{tabular}
  \label{tab:mathtemplate}
\end{table}

\section{Case Study}

\begin{tcolorbox}[
    title=Case Example 1,
    breakable,
    colback=white,
    boxrule=0.5pt,
    arc=2mm,
    left=1em,
    right=1em,
    top=0.8em,
    bottom=0.8em,
    fonttitle=\bfseries
]
\textbf{Problem:} In the school, there are 50 teachers, of whom 29 drink coffee, 28 drink tea, and 16 do not drink either coffee or tea. How many teachers drink only coffee, and how many drink only tea?

\vspace{0.5em}

\textbf{Long-CoT:} (Partial)\\
\hspace*{1em}\textbackslash nOkay, let's see. There's a problem about teachers who drink coffee or tea, and some who don't drink either. I need to find out how many drink only coffee and how many drink only tea. Hmm, this sounds like a problem where I can use sets and maybe a Venn diagram. Let me start by recalling some concepts.\textbackslash n\textbackslash nFirst, the total number of teachers is 50. Out of these, 29 drink coffee, 28 drink tea, and 16 don't drink either. \textbf{Wait}, so the number of teachers who drink at least one of the beverages should be the total minus those who don't drink either. Let me calculate that.\textbackslash n\textbackslash nTotal teachers = 50\textbackslash nTeachers who don't drink coffee or tea = 16\textbackslash nSo, teachers who drink coffee or tea or both = 50 - 16 = 34.\textbackslash n\textbackslash nOkay, so 34 teachers are in the union of coffee drinkers and tea drinkers. Now, I remember that the principle of inclusion-exclusion says that the number of people in either set is equal to the number in the first set plus the number in the second set minus the number in both sets. In formula terms:\textbackslash n\textbackslash nNumber of coffee drinkers + Number of tea drinkers - Number of both = Total in union\textbackslash n\textbackslash nPlugging in the numbers I have:\textbackslash n\textbackslash n29 (coffee) + 28 (tea) - Both = 34\textbackslash n\textbackslash nSo, 29 + 28 = 57. \whred{1173 characters}

\vspace{0.3em}

\textbf{R1-Compress:} (Partial)\\ 
\hspace*{1em}I need to find out how many teachers drink only coffee and how many drink only tea. This sounds like a problem where I can use sets and maybe a Venn diagram. The total number of teachers is 50. Out of these, 29 drink coffee, 28 drink tea, and 16 don't drink either. \textbf{Hmm}, the number of teachers who drink at least one of the beverages should be the total minus those who don't drink either. Let me calculate that.\textbackslash n\textbackslash nTotal teachers = 50\textbackslash nTeachers who don't drink coffee or tea = 16\textbackslash nSo, teachers who drink coffee or tea or both = 50 - 16 = 34.\textbackslash n\textbackslash nNow, I recall the principle of inclusion-exclusion: \textbackslash nNumber of coffee drinkers + Number of tea drinkers - Number of both = Total in union\textbackslash n\textbackslash nPlugging in the numbers:\textbackslash n\textbackslash n29 (coffee) + 28 (tea) - Both = 34\textbackslash n\textbackslash nSo, 29 + 28 = 57. \textcolor{lightgreen}{758 characters}

\end{tcolorbox}

As shown in Case 1, \methodname is able to successfully compress nearly 30\% of the content, while preserving the key computational steps and effectively retaining the reflection component (with “Wait” transformed into “Hmm”). Further analysis shows that the compressed content primarily consists of unnecessary discussions and redundant explanations.

\begin{tcolorbox}[
    title=Case Example 2,
    breakable,
    colback=white,
    boxrule=0.5pt,
    arc=2mm,
    left=1em,
    right=1em,
    top=0.8em,
    bottom=0.8em,
    fonttitle=\bfseries
]
\textbf{Problem:} Two is $10 \%$ of $x$ and $20 \%$ of $y$. What is $x - y$?\[
(\mathrm{A})\ 1 \qquad (\mathrm{B})\ 2 \qquad (\mathrm{C})\ 5 \qquad (\mathrm{D})\ 10 \qquad (\mathrm{E})\ 20
\]

\vspace{0.5em}

\textbf{Long-CoT:} (Partial)\\
\hspace*{1em}\textbackslash nOkay, let's see here. The problem says that two is 10\% of x and 20\% of y. Then we need to find x minus y. Hmm, percentages can sometimes trip me up, but let me take it step by step.\textbackslash n\textbackslash nFirst, let's parse the problem. Two is 10\% of x. So, in math terms, that would be 2 = 10\% of x. Similarly, two is also 20\% of y, so 2 = 20\% of y. Then, after finding x and y, subtract them to get x - y. The answer choices are numbers from 1 to 20, so it's probably a straightforward calculation once we figure out x and y.\textbackslash n\textbackslash nAlright, starting with the first equation: 2 is 10\% of x. To translate percentage to decimal, 10\% is 0.1. So, 2 = 0.1 * x. \whred{630 characters}

......

Wait, but let me make sure I didn't make any mistakes here. It's easy to mix up percentages. Let me go through the steps again just to confirm.\textbackslash n\textbackslash nFirst equation: 10\% of x is 2. 10\% is 0.1. So, 0.1x = 2. Solving for x: x = 2 \/ 0.1 = 20. Correct.\textbackslash n\textbackslash nSecond equation: 20\% of y is 2. 20\% is 0.2. So, 0.2y = 2. Solving for y: y = 2 \/ 0.2 = 10. Correct.\textbackslash n\textbackslash nSubtracting them: 20 - 10 = 10. Answer choice D. Hmm, seems solid. But the answer choices include 10, which is D, but I wonder if there's a trick here. Sometimes problems might involve something like the percentage of x equals the percentage of y but in different terms. Wait, no, in this case, it's straightforward: two is both 10\% of x and 20\% of y. So, x and y are different variables, each solved independently. Then subtract. Yeah, that's right.\textbackslash n\textbackslash nAlternatively, maybe someone might get confused and think it's the same number, but no, the problem clearly states two is 10\% of x and 20\% of y. So, x and y have different percentages leading to the same value, 2. Hence, solving for x and y gives different numbers, and their difference is 10.\textbackslash n\textbackslash nI don't see any mistakes here. Therefore, the answer should be D) 10.\textbackslash n\textbackslash n**Final Answer**\textbackslash n\boxed{D}\textbackslash n \whred{1193 characters}

\vspace{0.3em}

\textbf{R1-Compress:} (Partial)\\ 
\hspace*{1em}Okay, let's see here. Two is 10\% of x and 20\% of y. We need to find x minus y. Hmm, percentages can trip me up, so I'll take it step by step.\textbackslash n\textbackslash nFirst, let's parse the problem. Two is 10\% of x, so 2 = 10\% of x. Similarly, two is 20\% of y, so 2 = 20\% of y. Then, we need to find x and y, and subtract them to get x - y.\textbackslash n\textbackslash nStarting with the first equation: 2 is 10\% of x. 10\% is 0.1, so 2 = 0.1 * x. \textcolor{lightgreen}{394 characters}

......

Wait, let me recheck my steps. First, 10\% of x is 2, so 0.1x = 2 and x = 2 \/ 0.1 = 20. Correct.\textbackslash n\textbackslash nNext, 20\% of y is 2, so 0.2y = 2 and y = 2 \/ 0.2 = 10. Correct.\textbackslash n\textbackslash nSubtracting them: 20 - 10 = 10. Answer choice D. Hmm, seems solid. But I wonder if there's a trick. Wait, no, two is both 10\% of x and 20\% of y, so x and y are different variables. Alternatively, maybe someone might get confused and think it's the same number, but the problem clearly states two is 10\% of x and 20\% of y, so x and y have different percentages leading to the same value, 2. Hence, solving for x and y gives different numbers, and their difference is 10.\textbackslash n\textbackslash nI don't see any mistakes. So, the answer is D) 10.\textbackslash n\textbackslash n**Final Answer**\textbackslash n\boxed{D} \textcolor{lightgreen}{713 characters}

\end{tcolorbox}

As shown in Case 2, \methodname is able to successfully compress 236 characters during the problem-solving stage, while retaining the key steps and calculations. Furthermore, it effectively preserves the reflection and verification phases, compressing approximately 500 characters in this stage.

Overall, \methodname achieves high-quality compressed CoT by reducing redundant expressions and calculations, while maintaining the essential computations and reflective verification.

\section{Limitation}
\label{apd:limitaion}
Due to the inherent randomness in LLM-based compression—stemming from variations in prompts, different LLMs, and sampling parameters—the quality of each compressed instance cannot be fully guaranteed. Although we generate multiple candidate chunks and apply a search strategy to select high-quality compressed CoT, it is still possible to obtain outputs with contextual incoherence. This work provides insights into Long-CoT compression from the perspective of reflection and, through case studies, reveals that the removed tokens are primarily associated with redundant expressions and repetitive computational steps. Nonetheless, the compression of Long-CoT remains an open problem and warrants further investigation.

\end{document}